\documentclass[runningheads,a4paper]{llncs}

\usepackage{amsmath}
\usepackage{url}          
\usepackage[pdftex]{graphicx}
\usepackage{fullpage}

\begin{document}

\title{Image classification and retrieval with random depthwise signed convolutional neural networks}

\author{Yunzhe Xue \and Usman Roshan}
%
%
%
\institute{Department of Computer Science, New Jersey Institute of Technology, Newark NJ 07090, USA,\\
\email{yx277@njit.edu,usman@njit.edu}
}

\maketitle

%

%

\begin{abstract}
We propose a random convolutional neural network to generate a feature space in which we study image classification and retrieval performance. Put briefly we apply random convolutional blocks followed by global average pooling to generate a new feature, and we repeat this $k$ times to produce a $k$-dimensional feature space. This can be interpreted as partitioning the space of image patches with random hyperplanes which we formalize as a random depthwise convolutional neural network. In the network's final layer we perform image classification and retrieval with the linear support vector machine and $k$-nearest neighbor classifiers and study other empirical properties. We show that the ratio of image pixel distribution similarity across classes to within classes is higher in our network's final layer compared to the input space. When we apply the linear support vector machine for image classification we see that the accuracy is higher than if we were to train just the final layer of VGG16, ResNet18, and DenseNet40 with random weights. In the same setting we compare it to a recent unsupervised feature learning method and find our accuracy to be comparable on CIFAR10 but higher on CIFAR100 and STL10. We see that the accuracy is not far behind that of trained networks, particularly in the top-$k$ setting. For example the top-2 accuracy of our network is near 90\% on both CIFAR10 and a 10-class mini ImageNet, and 85\% on STL10. We find that $k$-nearest neighbor gives a comparable precision on the Corel Princeton Image Similarity Benchmark than if we were to use the final layer of trained networks. As with other networks we find that our network fails to a black box attack even though we lack a gradient and use the sign activation. We highlight sensitivity of our network to background as a potential pitfall and an advantage. Overall our work pushes the boundary of what can be achieved with random weights.
\end{abstract}


\section{Introduction}
Convolutional neural networks (CNNs) are the state of the art in image recognition
benchmarks today \cite{NIPS2012_4824}. Optimization methods such as
stochastic gradient descent \cite{bottou2010large} combined with data augmentation 
\cite{salamon2017deep}, regularization \cite{le2013building}, dropout and cutout \cite{srivastava2014dropout,devries2017improved}
have made CNNs the de-facto approach for accurate image recognition \cite{krizhevsky2012imagenet}.
Interestingly, several of these methods involve randomness. 

For example the dropout method \cite{srivastava2014dropout} ignores a random set of nodes during training. 
The cutout method  \cite{devries2017improved} masks random square 
patches in the input training images. Stochastic gradient
descent randomly uses a single training example at a time (or mini batches) to obtain the 
gradient as opposed to computing it from the entire dataset. This method has been
shown to converge to the global optimum for convex functions as we increase
the number of iterations \cite{murata1998statistical}. All of these methods use randomness
to avoid overfitting during training and thus give better model generalization.

Random weights have been explored previously
in several  studies \cite{saxe2011random,jarrett2009best,pinto2009high,gilbert2017towards}
including generating images with random nets \cite{he2016powerful}. 
They show the importance of a network architecture in achieving high accuracies and connect unsupervised
pre-training and discriminative fine tuning to architecture. We explore randomness from the perspective of 
random depthwise convolutional blocks. 

Consider a series of convolutional blocks applied repeatedly to an image. The image representation in the
final layer is then globally average pooled to obtain a single value. If we repeat this $k$ times we obtain
a $k$ dimensional space. This can also be considered as a method for unsupervised feature learning
since we make no use of labels in generating the new space. We show below that our method can be 
interpreted as applying random hyperplanes to all patches of all input images and average pooling the final image
representations. We formalize this as a random depthwise convolutional neural network and study various aspects of 
image classification and retrieval in our network's final layer. 

We present several experimental results on image classification and retrieval in our network's final layer. We
start by showing that  images across and within classes are better represented than the input feature space. We then
compare to the final trained layer of other random networks and an unsupervised feature learning method. In both cases our network attains comparable or higher accuracies. Compared to trained networks our random weights are no match but are not far behind in accuracy, especially as we go into top-2 and top-3 accuracy. 

Interestingly our network performs competitively to trained networks when it comes to the problem of image retrieval, particularly image retrieval by similarity such as the Corel-Princeton Image Similarity benchmark \cite{corel}. Finally we highlight some limitations sensitivity of our network's layer to background colors which can be a disadvantage but possibly also an advantage in some cases. Overall we push the envelope of random networks and show that accuracies better than
expected can be achieved. 



\section{Methods}
\subsection{Motivation and intuition behind depthwise random convolutions}
\begin{figure}[h]
  \centering
  \setlength\tabcolsep{12pt} 
  \begin{tabular}{ccc}
\includegraphics[scale=.3]{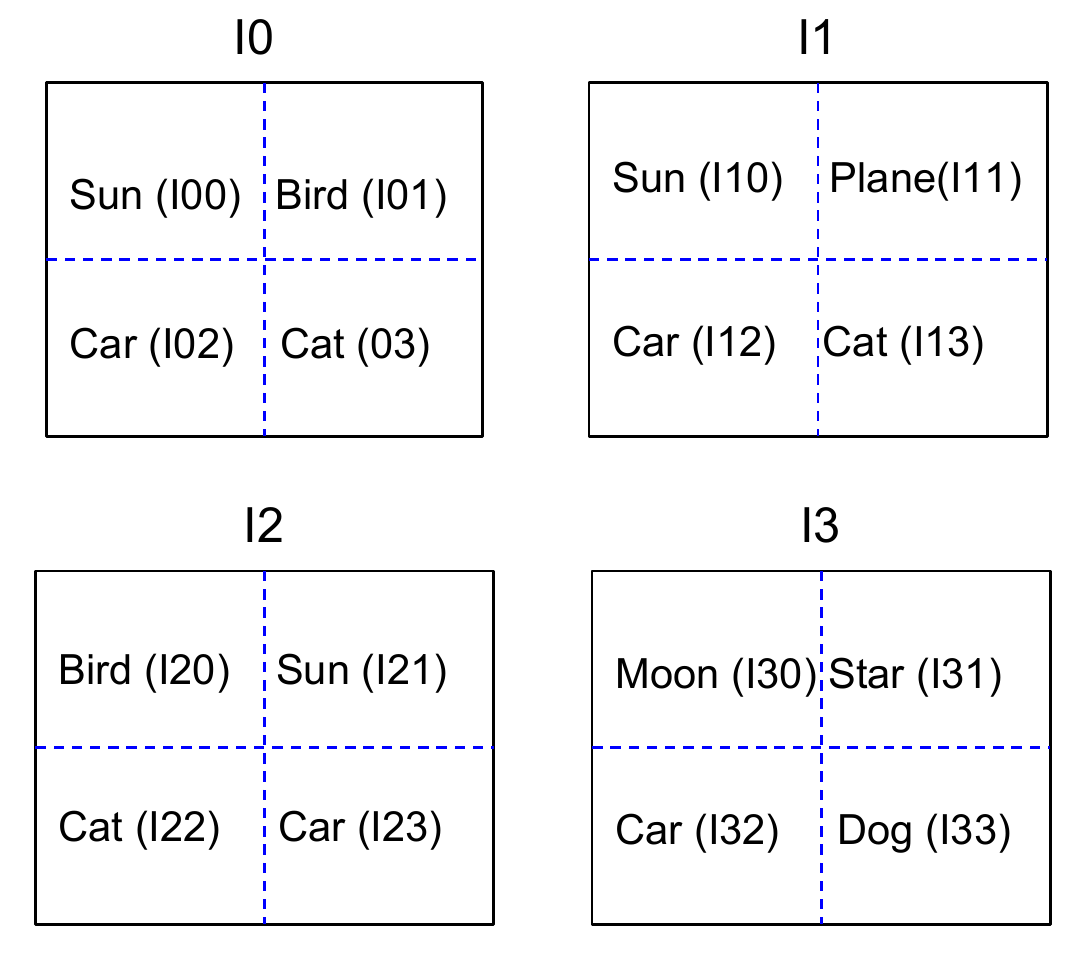} &  \includegraphics[scale=.25]{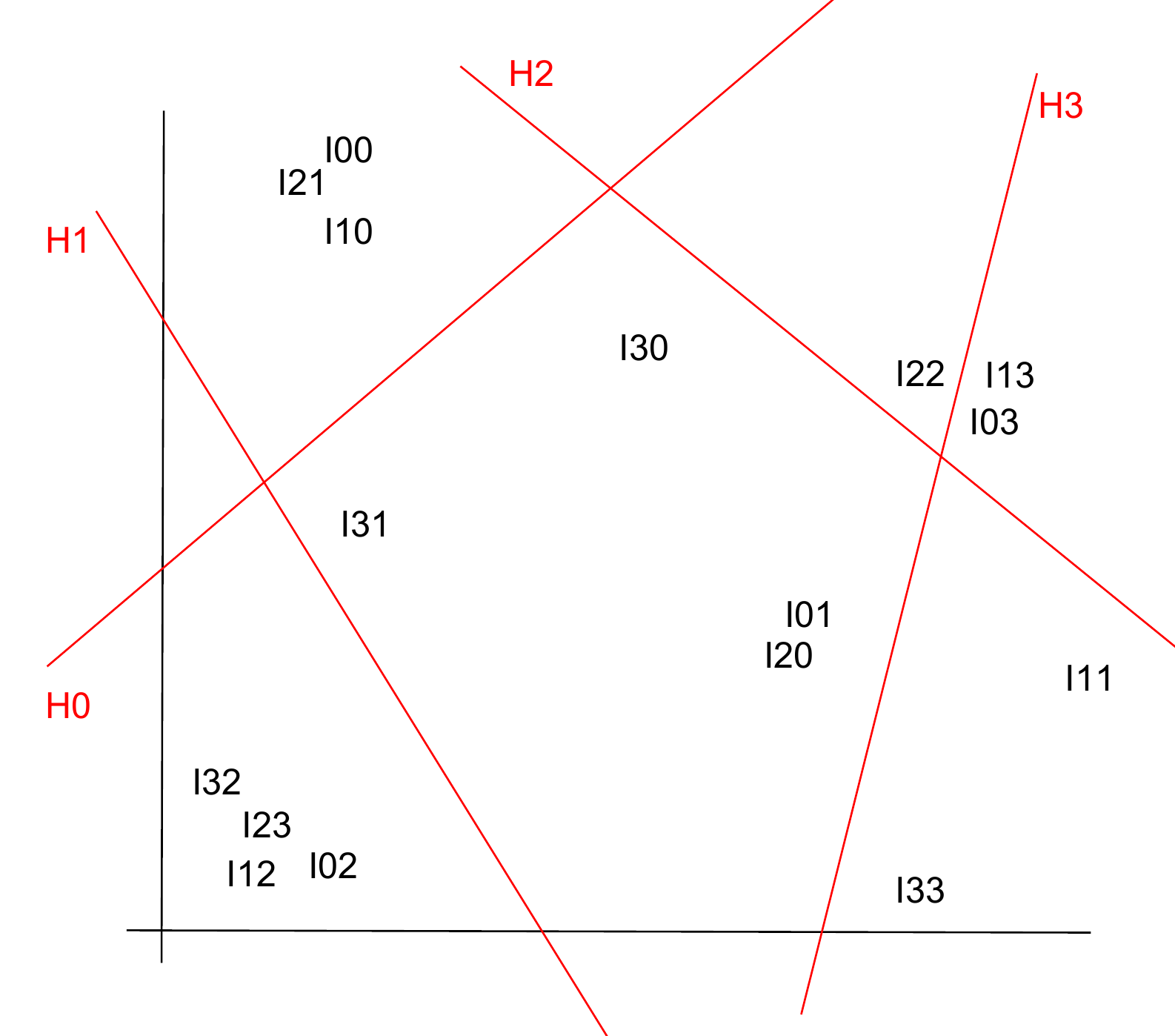}  &   \includegraphics[scale=.35]{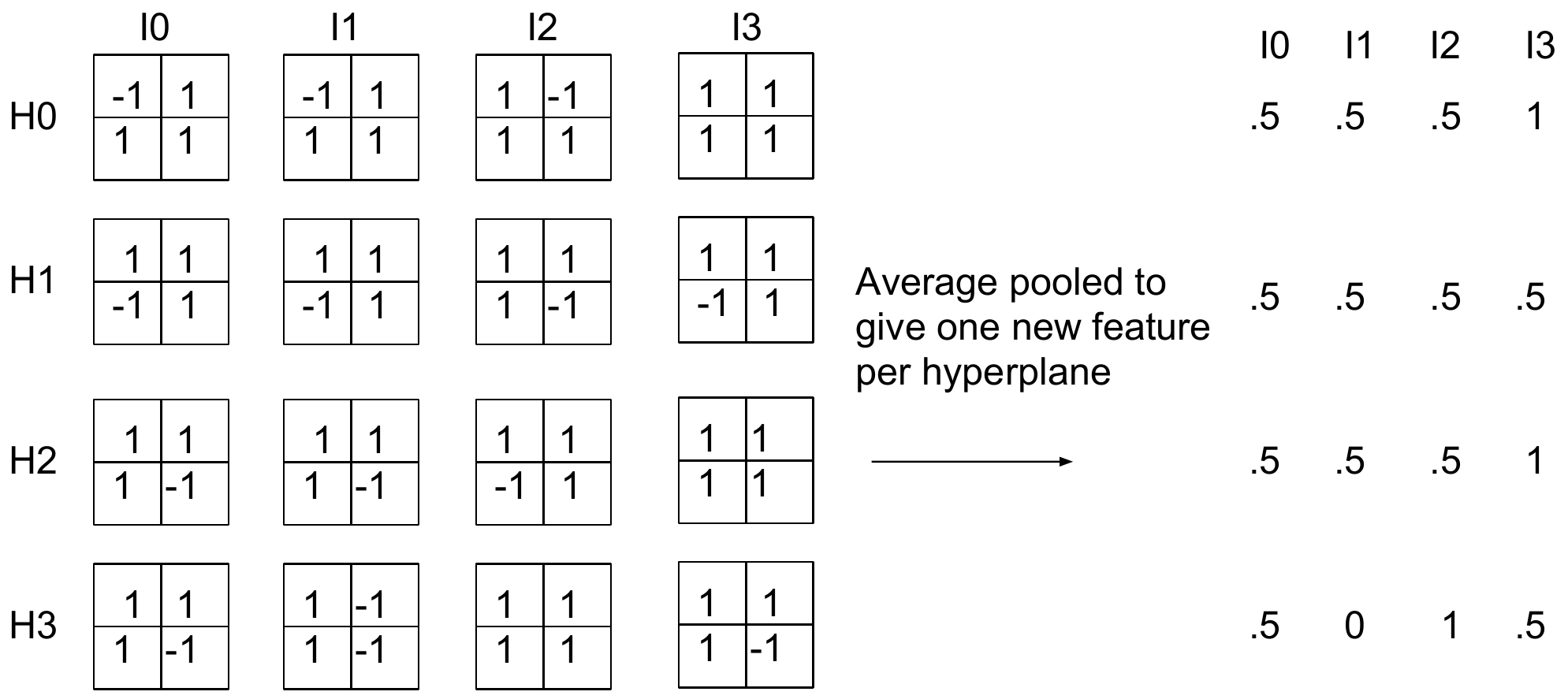} \\
(a) & (b) & (c) \\
\end{tabular}
  \caption{Shown in (a) are four images each containing objects in different parts of the image and divided into four partitions. In (b) we show four random hyperplanes (in red) on the input space of features from all patches of the images. The outputs of the four random hyperplanes correspond to a convolutional kernel that considers just four partitions (part (c)). We average pool to obtain a feature value. \label{nn2}}
\end{figure}

We provide motivation and intuition with a simple toy example in Figure~\ref{nn2}. There we see four images
$I_0, I_1, I_2$, and $I_3$ containing various objects. Clearly images $I_0$ and $I_1$ are very similar to each 
other. Image $I_2$ is also similar since it contains common objects as $I_0$ and $I_1$ but they are in different
positions. Image $I_3$ shares only one common object to the other images and thus is the most dissimilar. We
seek a representation that would capture these similarities. 

Suppose we divide each image into four equal quadrants (or patches) and consider four random hyperplanes $H_0,H_1,H_2$, and $H_3$ in the space of all  patches as shown in Figure~\ref{nn2}. We place patches containing the same object nearby in the figure since they are likely to be similar in pixel values. For example the car in partitions
$I_{02},I_{12},I_{23}$, and $I_{32}$ are clustered in the lower left in the middle figure near the origin.

We determine the sign of each patch according to each hyperplane and obtain a $2\times 2$ matrix for each image. This is exactly the output of a convolution kernel that considers just four partitions of an image. We then average pool the matrix to obtain a single feature value for the image given by the hyperplane. By repeating this for many hyperplanes we obtain more features per image. We see in our toy example that the final feature representation for images $I_0,I_1$, and $I_2$ are more similar than image $I_3$. 

We don't have a theoretical guarantee that applying random hyperplanes to image patches repeatedly
(as we do in our network) would yield a linearly separable space. However, one can draw intuition from our
toy example in Figure~\ref{nn2}. There we see that the four random hyperplanes partition the space and
that patches in the same space will have exactly the same outputs for all four hyperplanes. 
Patches in adjacent partitions are likely to be  less similar and will have exactly one of the four hyperplane 
outputs to be different.



\subsection{Random depthwise convolutional neural networks}
Before formalizing our notion into random depthwise networks we briefly review convolutional 
neural networks. Convolutional neural networks are typically composed of alternating convolution and pooling layers
followed by a final flattened layer. A convolution layer is specified by a kernel size and the number of
kernels in the layer. Briefly, the convolution layer performs a a moving non-linearized dot product against pixels given by a 
fixed kernel size $k \times k$ (usually $3\times3$ or $5\times5$). The dot product is usually
non-linearized with the sigmoid or hinge (relu) function since both are differentiable and fit into
the gradient descent framework. The output of applying a
$k\times k$ convolution against a $p\times p$ image is an image of size $(p-k+1) \times (p-k+1)$. 


Consider applying random convolutional blocks repeatedly and then averaging all the values in the final
representation of the image. If we repeat this $k$ times it gives us $k$ new features. This can
be described as a random depthwise convolutional neural network (RDCNN). Each convolutional block in our network is
a convolutional kernel followed by $2\times2$ average pooling with stride 2. 

\begin{figure}
  \centering
\includegraphics[scale=.4]{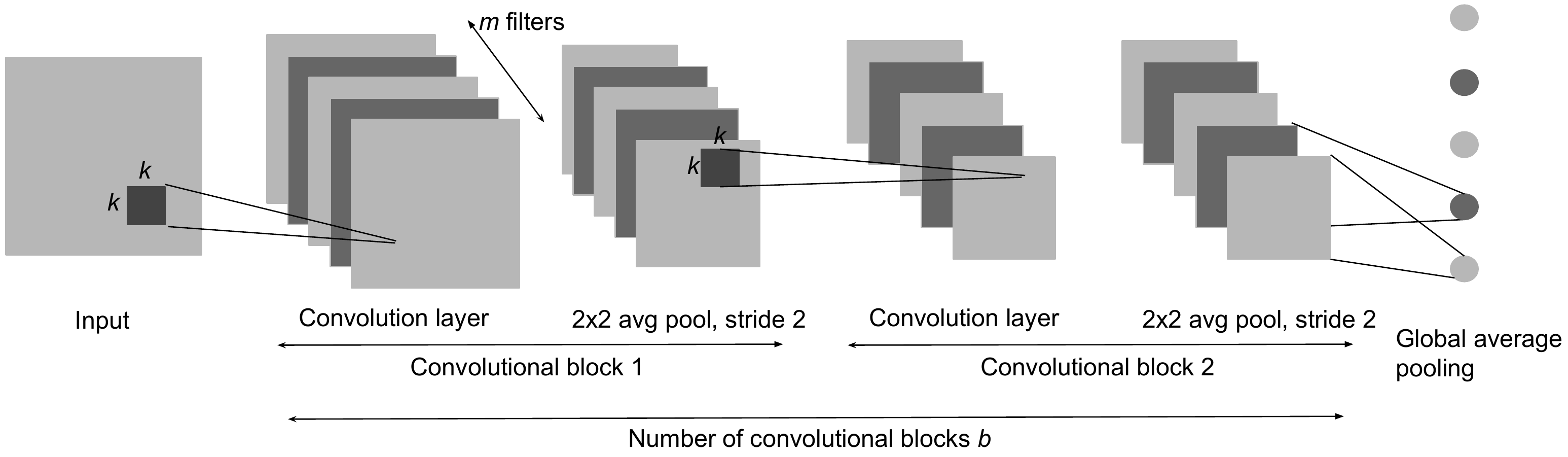} 
  \caption{A random depthwise convolutional neural network with two convolutional blocks, kernel size of $k$, and
  $m=5$ kernels in each layer \label{pcnn}}
\end{figure}

Our network is parameterized by the number of convolutional blocks $b$, the size of each kernel $k \times k$ and the 
number of kernels $m$ in each layer (this is the same in each layer). In Figure~\ref{pcnn} 
we show an example of our network with two layers ($l=2$) and five $3\times 3$ 
convolution blocks in each layer ($m=5,k=3$). We set the values in each convolutional kernel randomly from the
Normal distribution with mean 0 and variance 1.

We non-linearize the output of each convolution with the sign function and our convolution 
is \emph{depthwise}. This means the $i^{th}$ convolution is applied on the $i^{th}$ kernel only of the
previous layer. In the input layer, however, the convolution is applied in the conventional way to account
for RGB images that have three layers. After we are done with convolutions 
we globally average pool the final layer which gives us a flattened feature space. We then 
apply a linear support vector machine or stochastic gradient descent on the final feature space.

\subsection{Experimental performance study}
In order to evaluate the empirical performance of our random network we compare it to three
popular networks with random and trained weights and an unsupervised feature learning method 
on several image benchmarks.

\subsubsection{Deep networks compared in our study}
We compare our method to modern networks used in image recognition today. These are all
convolutional neural networks designed to enable deeper architectures and are trained with 
stochastic gradient descent. We implement these networks ourselves with PyTorch and TensorFlow and 
make our implementations freely available from the study website
\url{https://github.com/xyzacademic/RandomDepthwiseCNN}.

\begin{itemize}
\item ResNet18 \cite{he2016deep}: Residual convolutional networks contain connections from previous layers and not just the last one.
\item DensenNet40 \cite{huang2017densely}: Convolutional networks contain dense layers in between convolutions.
\item VGG16 \cite{simonyan2014very}: Deep convolutional neural network with layers of convolution and pooling.
\end{itemize}

\subsubsection{Datasets}
We collect several image benchmarks on which we evaluate our method.

\begin{itemize}
\item MNIST \cite{lecun1998gradient}: Handwritten digit recognition from 10 classes in $32\times 32$ images, training size of 60,000 and test size of 10,000
\item CIFAR10 \cite{krizhevsky2009learning}: Object recognition from 10 classes in $32\times 32$ color images, training size of 50,000, test size of 10,000
\item CIFAR100 \cite{krizhevsky2009learning}: As CIFAR10 except from 100 classes 
\item STL10 \cite{coates2011analysis}: Object recognition from 10 classes in $96\times96$ color images, training size of 5000, and test size of 8000
\item Mini-ImageNet \cite{ILSVRC15}: We randomly select 10 classes from the benchmark giving a total of 
12,730 training and 500 test color images each of size $256\times256$. 
We provide these set of images on the study's website at
\url{https://github.com/xyzacademic/RandomDepthwiseCNN}.
\item Corel Princeton Image Similarity \cite{corel}: Eight query images and their similar images (totaling 10,000) ranked by humans. We provide the image benchmark on our study's website \url{https://github.com/xyzacademic/RandomDepthwiseCNN}.
\end{itemize}

\subsubsection{Experimental platform and source code}
We conduct all experiments on computing nodes equipped with
Intel Xeon E5-2630-v4 CPUs and NVIDIA Tesla P100 16GB Pascal GPUs.
We implement our method to produce the final flattened layer with PyTorch and TensorFlow and make it 
available on this study's website \url{https://github.com/xyzacademic/RandomDepthwiseCNN}. 
We also provide there our implementations of other networks that we study in this paper including code
to generate adversarial examples.

\subsubsection{Program parameters and training}
We use the libinear program \cite{liblinear} version 2.20 for determining a linear support
vector machine on the final layer obtained by our model. Our liblinear parameters are -s 2 -B 1
which turns on primal optimization and a threshold value of non zero. 
Our $C$ (regularization) parameter values are 0.5 for CIFAR10, CIFAR100, and STL10, and 0.01 for MNIST and Mini-ImageNet. We optimize ResNet18, DenseNet40, and VGG16 networks with stochastic gradient descent. For
CIFAR10 and CIFAR100 we use a batch size of 256 whereas for STL10 and Mini-ImageNet we use 32 and 16
respectively. On Mini-ImageNet we use 90 epochs and on the other benchmarks we use 300. We vary the learning
rate across the number of epochs by starting with a large value of 0.1 and progressively reducing to 0.01 and 0.001 as 
the number of epochs increases.

\section{Results}

\subsection{Effect of number of blocks and kernel; size}
We start by exploring the effect of number of blocks and kernel size on our model test accuracy. 
In Table~\ref{blockslayers} we show the accuracy of our model with a fixed
number of layers, fixed kernel size, and 100,000 kernels. We see that the number of layers and
kernel size has a considerable effect on test accuracy 

\begin{table}[h]
  \centering
  \begin{tabular}{ccccc}
  \hline
 Dataset  & $k=3,b=1$ & $k=3,b=2$ & $k=5,b=1$ &$ k=5,b=2$  \\
  \hline 
  CIFAR10 &   75.8  & 74.8 & 76.4 & 70.1  \\
  CIFAR100 & 51.8 & 51.9 & 53.3 & 47.5  \\
    \hline 

  \end{tabular}
  \caption{Accuracy of our network RDCNN with different number of convolutional blocks and kernel size. 
  \label{blockslayers}}
\end{table}

In Table~\ref{params} we show the parameters used in RDCNN for each image benchmark. These parameters give
us the best test accuracy over combinations of $k=3,5,7$ and $b=1, 2, 3$. 

\begin{table}[h]
  \centering
  \begin{tabular}{cccccc}
  \hline
 Dataset  & MNIST & CIFAR10 & CIFAR100& STL10& Mini ImageNet \\
  Parameters & $k=7,b=1$ & $k=5,b=1$ & $k=5,b=1$ & $k=3,b=3$, $k=5,b=2$ \\ \hline
  \end{tabular}
  \caption{Parameters used in RDCNN for each image benchmark  \label{params}}
\end{table}

\subsection{Effect of number of random kernels on test accuracy}
\begin{figure}[h]
  \centering
\includegraphics[scale=.5]{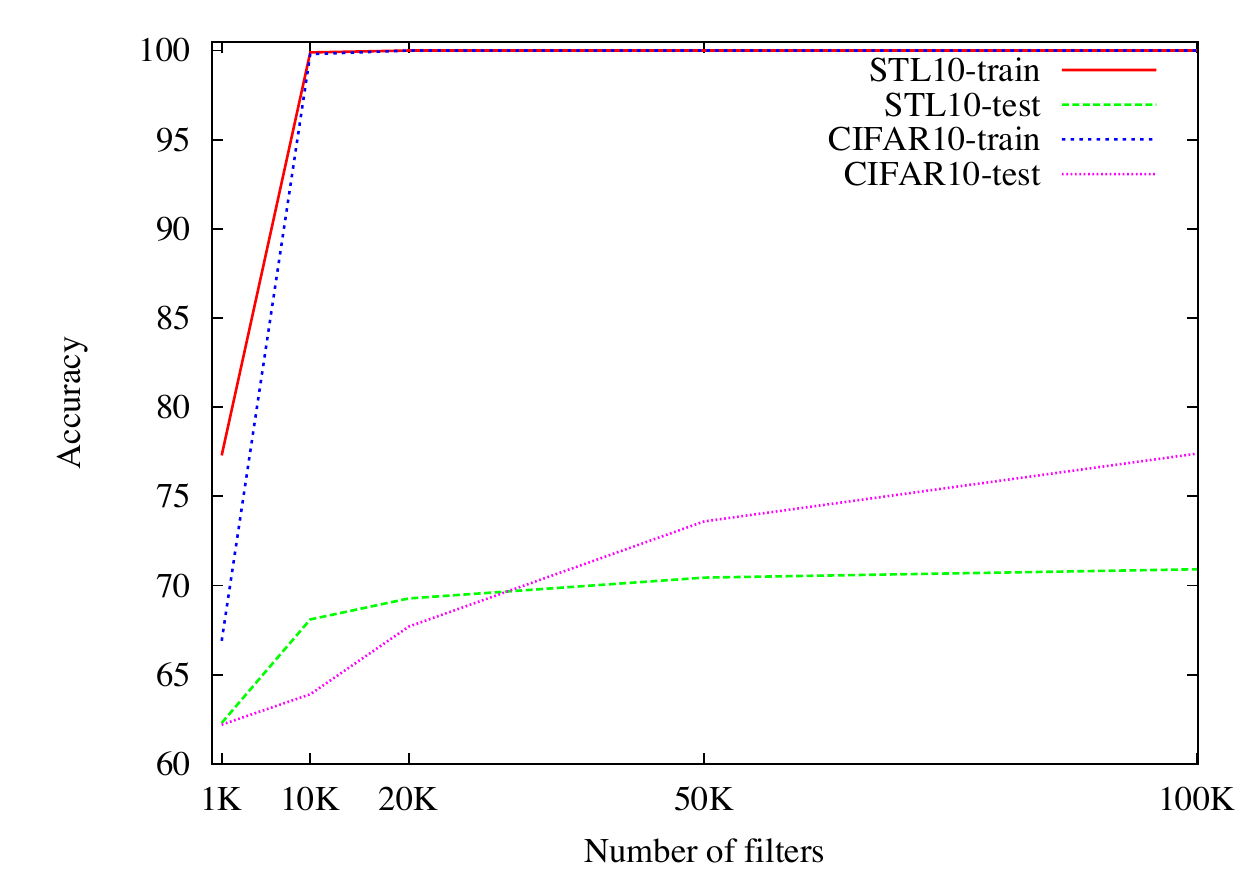} \\
  \caption{Effect of number of kernels (final features) on the test accuracy \label{features}}
\end{figure}

Having shown the effect of kernel size and number of convolutional blocks we proceed to determine
the effect of number of kernels on the test accuracy. Each kernel gives rise to a new feature in the final 
flattened layer. We expect an improvement in test accuracy as we increase the number of kernels
and indeed we see this is the case in Figure~\ref{features}. There we see that increasing the number of
kernels improves the accuracy on the STL10 and CIFAR10 benchmarks. We also see that
the train accuracy reaches 100\% much faster and stays there while test accuracy continues
to improve.

\subsection{Image pixel distribution similarity across and within classes}
\begin{figure}[h]
  \centering
  \begin{tabular}{cc}
\includegraphics[scale=.4]{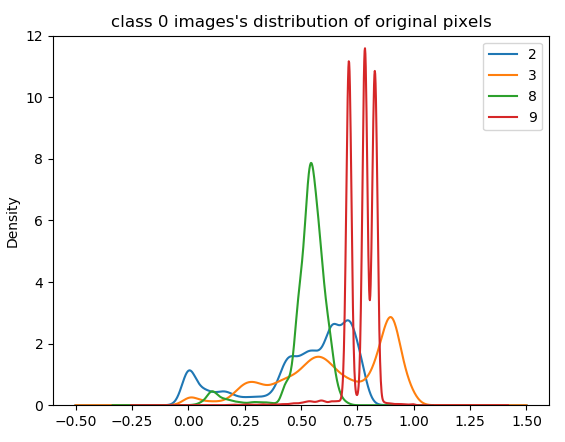} & \includegraphics[scale=.4]{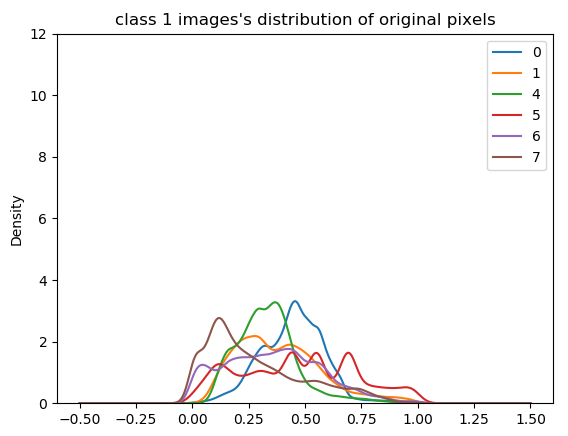} \\
\multicolumn{2}{c}{(a)} \\
\includegraphics[scale=.4]{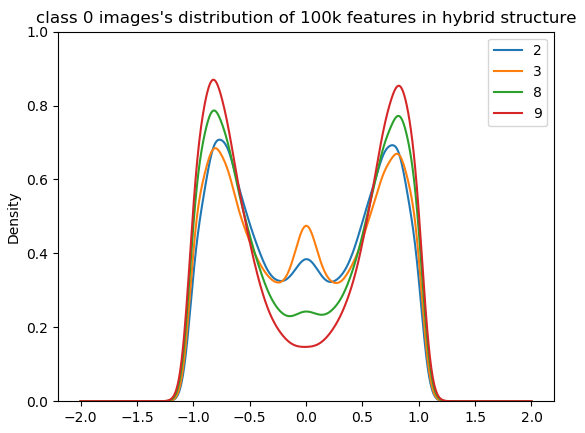} & \includegraphics[scale=.4]{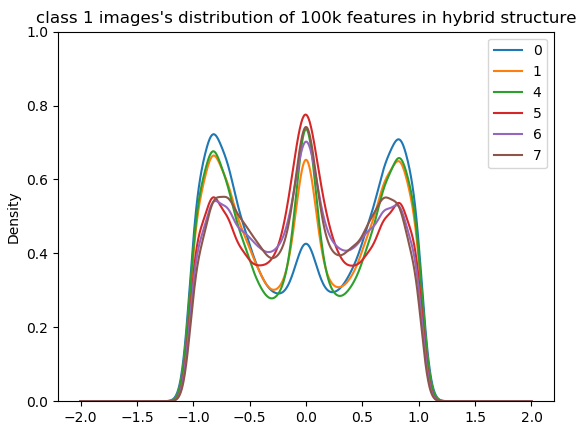} \\
\multicolumn{2}{c}{(b)} \\
\end{tabular}
  \caption{Pixel distribution of images from two classes in the original and new representations \label{fig1}}
 \end{figure}

We select 10 random images from STL10 (6 from class 0 and 4 from class 1) and show the distribution of the image pixel values before and after our network. In Figure~\ref{fig1}(a) we see the distribution of image pixel values in the original data and in (b) we show the distribution in our final layer before the SVM. We see that the random kernels smoothen the image distributions and appear to show a better separation of image distributions between the two classes. 

In order to obtain a quantitative measure we report the average Jensen Shannon (avgJS) divergence \cite{fuglede2004jensen} 
between image pixel distributions across classes and within classes. To obtain this we simply average the divergence across all pairs of image distributions across two classes (and within classes). We then measure the ratio of 

\begin{equation*}
\frac{avgJS(class0,class1)}{avgJS(class0,class0)+avgJS(class1,class1)}
\end{equation*}

For images in the original feature representation we find a ratio of $\frac{0.25}{0.3 + 0.18}=0.52$ and in our RDCNN final layer the ratio increases to $\frac{0.02}{0.018+0.016}=0.59$. This ratio varies across classes and the above values are for class 0 and 1 that we randomly chose. If we measure this ratio between classes 0 and 6 we see a larger difference in the ratio across the two feature representations. In the original representation this value is $\frac{0.24}{0.3+0.13}=0.56$ and in our RDCNN final layer it increases to $\frac{0.025}{0.018+0.015}=0.76$. Thus we see that our new feature representation gives a better \emph{signal to noise} ratio.

\subsection{Comparison to random weights in other networks and unsupervised feature learning}

\begin{figure}[h!]
  \centering
\includegraphics[scale=.5]{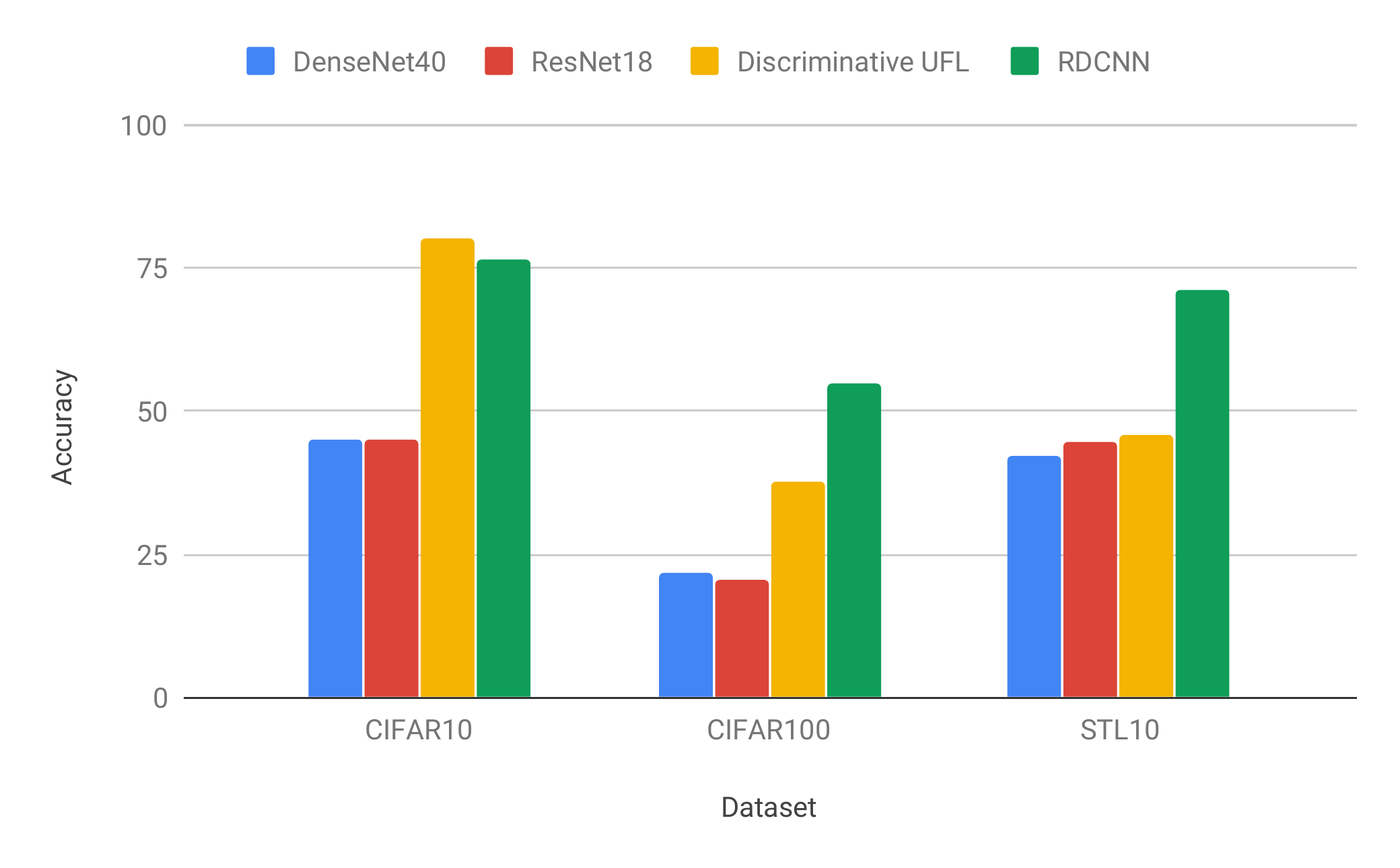} 
  \caption{Test accuracy of (1) deep networks DenseNet40 and ResNet18 with random weights except for training in final layer, (b) unsupervised feature learning with convolutional networks denoted as Discriminative UFL, and (3) our network RDCNN 
 on three image benchmarks \label{random}}
 \end{figure}

Here we consider random weights in ResNet18 and DenseNet40 and except for the final layer that we train. We also consider a recent unsupervised feature learning method (Discriminative UFL) \cite{wu2018unsupervised} that targets instance level discrimination with convolutional neural networks. In Figure~\ref{random} we see that all the random networks have lower accuracy than ours (denoted as RDCNN) even after training the final layer. We obtain the same accuracy with Discriminative UFL on CIFAR10 as in their published paper. However, when we ran their code on the other benchmarks the accuracies were much lower.

The accuracies of random networks on CIFAR10 that we report here are lower than those of previously reported by
Saxes et. al. \cite{saxe2011random} (53.2\%) and Gilbert el. al. \cite{gilbert2017towards} (74.8\%). It's likely they may have trained additional parameters besides just the final layer that we do. Jarrett et. al. \cite{jarrett2009best} report 99.46\% on MNIST with random weights and we perform comparably by reaching 99.4\%.


\subsection{Comparison to trained convolutional networks}
\begin{figure}[h]
  \centering
  \begin{tabular}{cc}
\includegraphics[scale=.3]{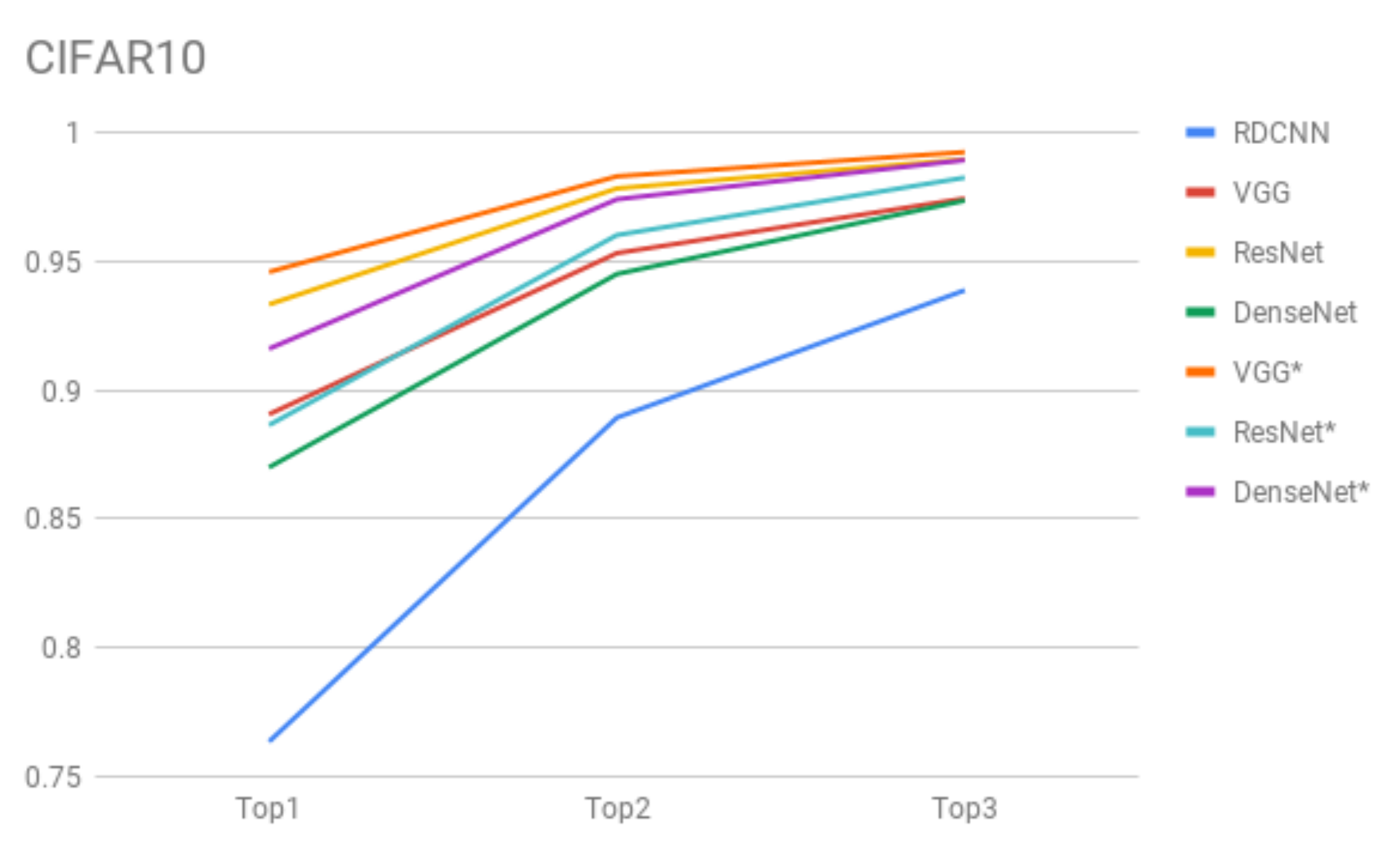} & \includegraphics[scale=.3]{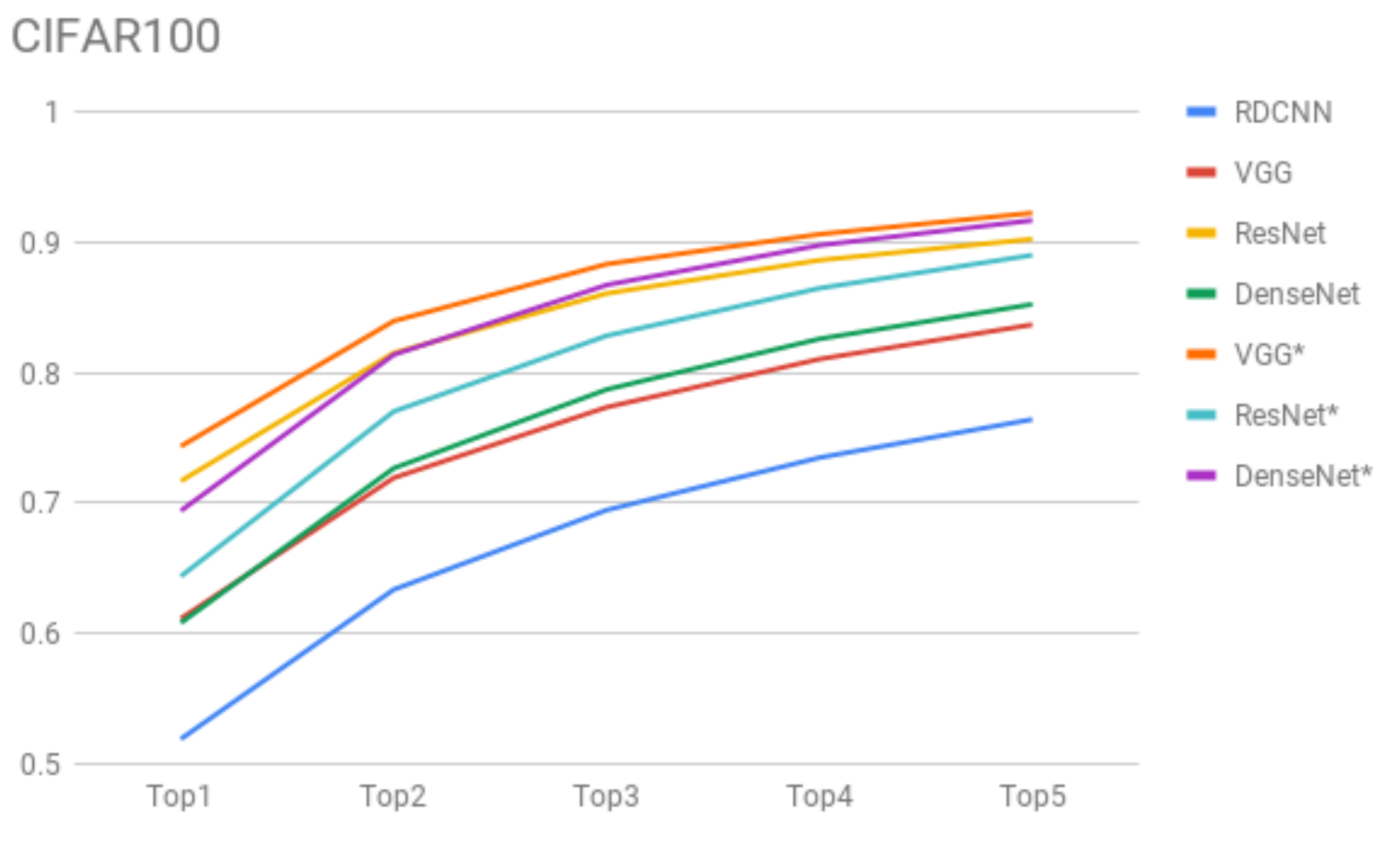} \\
\includegraphics[scale=.3]{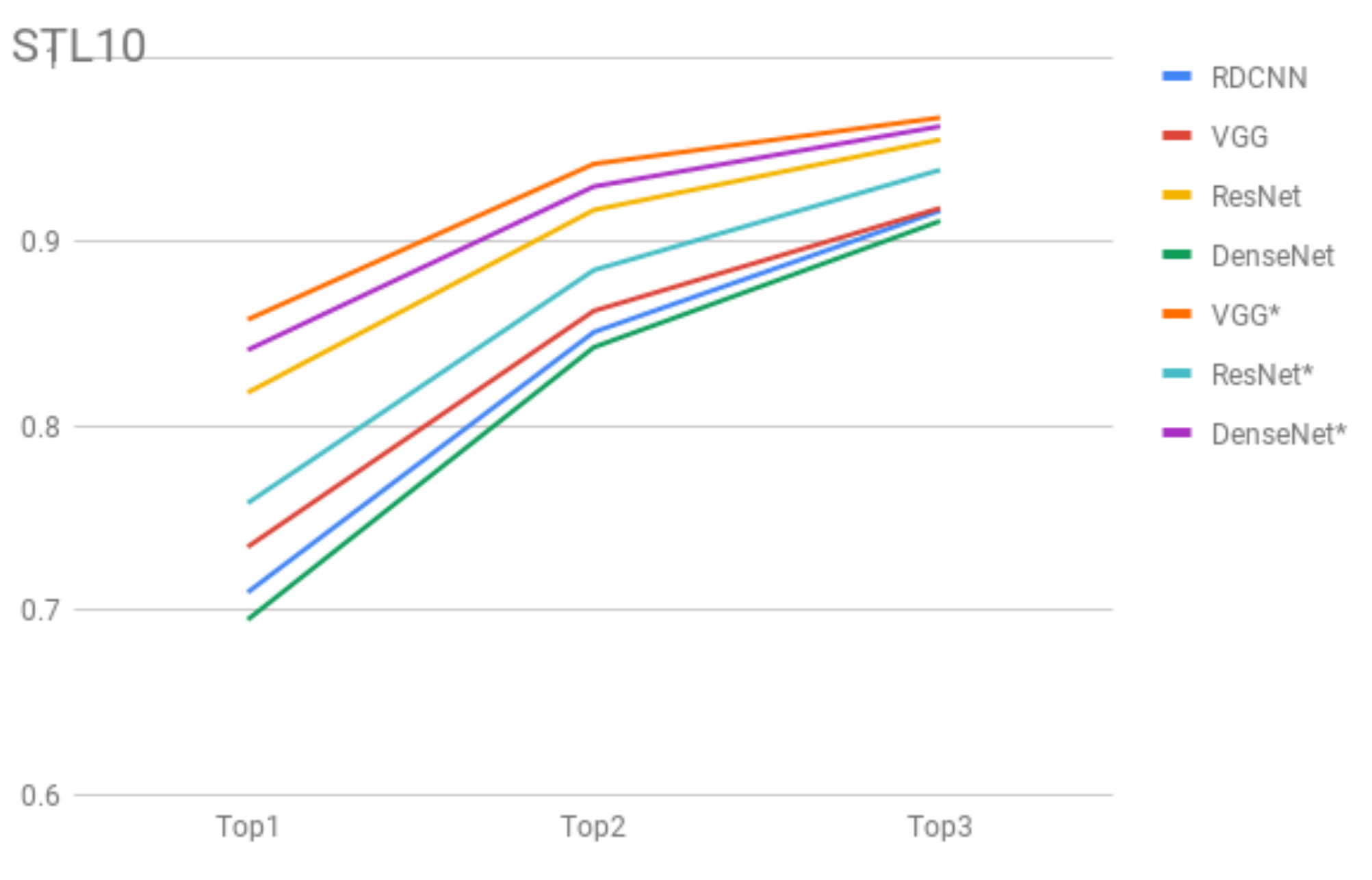} & \includegraphics[scale=.3]{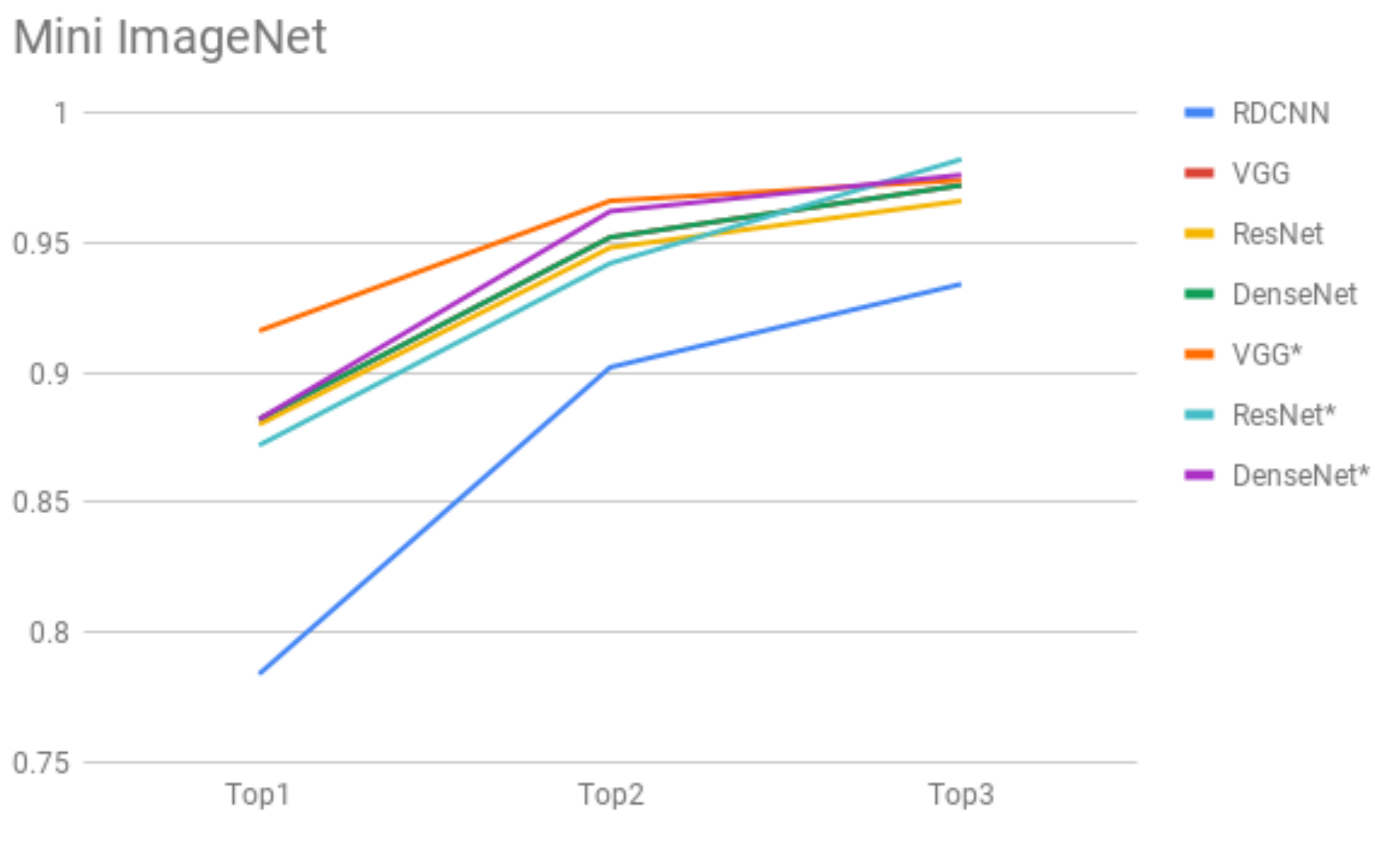} \\
\end{tabular}
  \caption{Top-k accuracy of different networks on benchmarks. Methods with asterix denote 
  data augmentation was enabled. \label{benchmarkacc}}
\end{figure}

In Figure~\ref{benchmarkacc} we show the test accuracies of the top-1, top-2, and top-3 outputs from
all networks on four benchmarks. The top-k outputs are obtained by considering images with 
the top $k$ highest outputs in the final layer. In our case we use the SVM discriminant to rank the
outputs. 

In STL10 our random network (RDCNN) performs comparably to trained
networks without data augmentation. In fact it performs better than when DenseNet40 is trained without
data augmentation. On the other benchmarks our network is trailing in accuracy
but that difference becomes smaller as we consider top-2 and top-3 outputs of the classifier. 

In both CIFAR10 and Mini ImageNet we see that the random network has the steepest increase
in accuracy from top-1 to top-2. In fact in top-2 our network is about 90\% accurate on CIFAR10
and Mini ImageNet and 85\% on STL10. In CIFAR100 that has a 100 classes our network catches 
up to trained networks at a much slower pace. 


\begin{figure}[h]
  \centering
  \begin{tabular}{c}
\includegraphics[scale=.3]{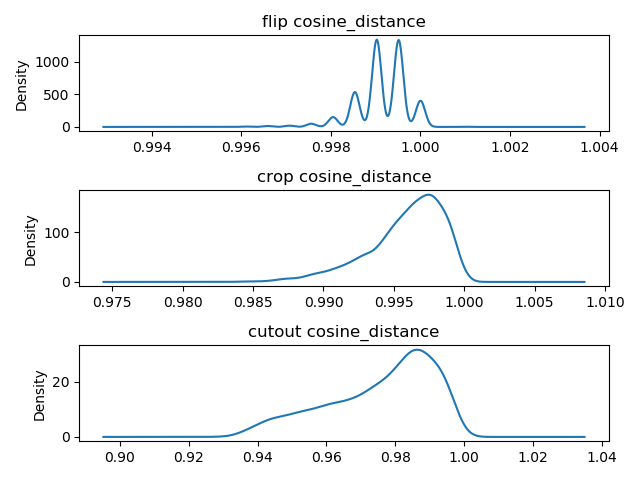} \\
\end{tabular}
  \caption{Cosine similarity distribution across all STL10 images. Each cosine value is between an image
  and its augmentation in our network's final layer feature space. \label{cos}}
\end{figure}  

\paragraph{Data augmentation} For our method we first separately generated augmented images 
with flips and random rotations from STL10 (10 augments per input image). We then combined these 
into the original training set and
used them as input to our network. Our final test accuracy, however, was no better than training on just the
original data. To understand this we compare the cosine similarity of each augmented image to its original and plot
the cosine similarity value distribution across all images. We perform the cosine similarity in the feature space
given by the final layer of our network. In Figure~\ref{cos} we see that the flips and
rotations produce images that are highly similar to the original in our feature representation. However, if we
perform the cutout augmentation \cite{devries2017improved} the images are relatively more different. Perhaps
this augmentation or something even stronger may boost the accuracy of our network.

\subsection{Similar image retrieval}
We train the VGG16 and ResNet18 networks on all 10,000 images in the Corel Princeton Image Similarity dataset \cite{corel}
except for the eight queries. In this benchmark the most similar images are ranked and given a score by
human observers. We use the last hidden layer for representing all images. Our choice 
for this comes from  previous studies that have shown the final hidden layer (that is usually a dense layer) 
works best for image retrieval with deep convolutional neural networks \cite{wang2015deep,sharif2014cnn,wan2014deep}. 

Here we use our network's final layer to represent images. We also study a version of our network where we 
perform training on the final layer. We use our final layer as input into a single layer multilayer perceptron with sigmoid activation 
and 1000 features in the hidden layer (as implemented in Python scikit-learn \cite{scikit}). 
Here we exclude the eight queries from the training procedure.

\begin{figure}[h]
  \centering
  \setlength\tabcolsep{4pt} 
  \begin{tabular}{cc}
\includegraphics[scale=.17]{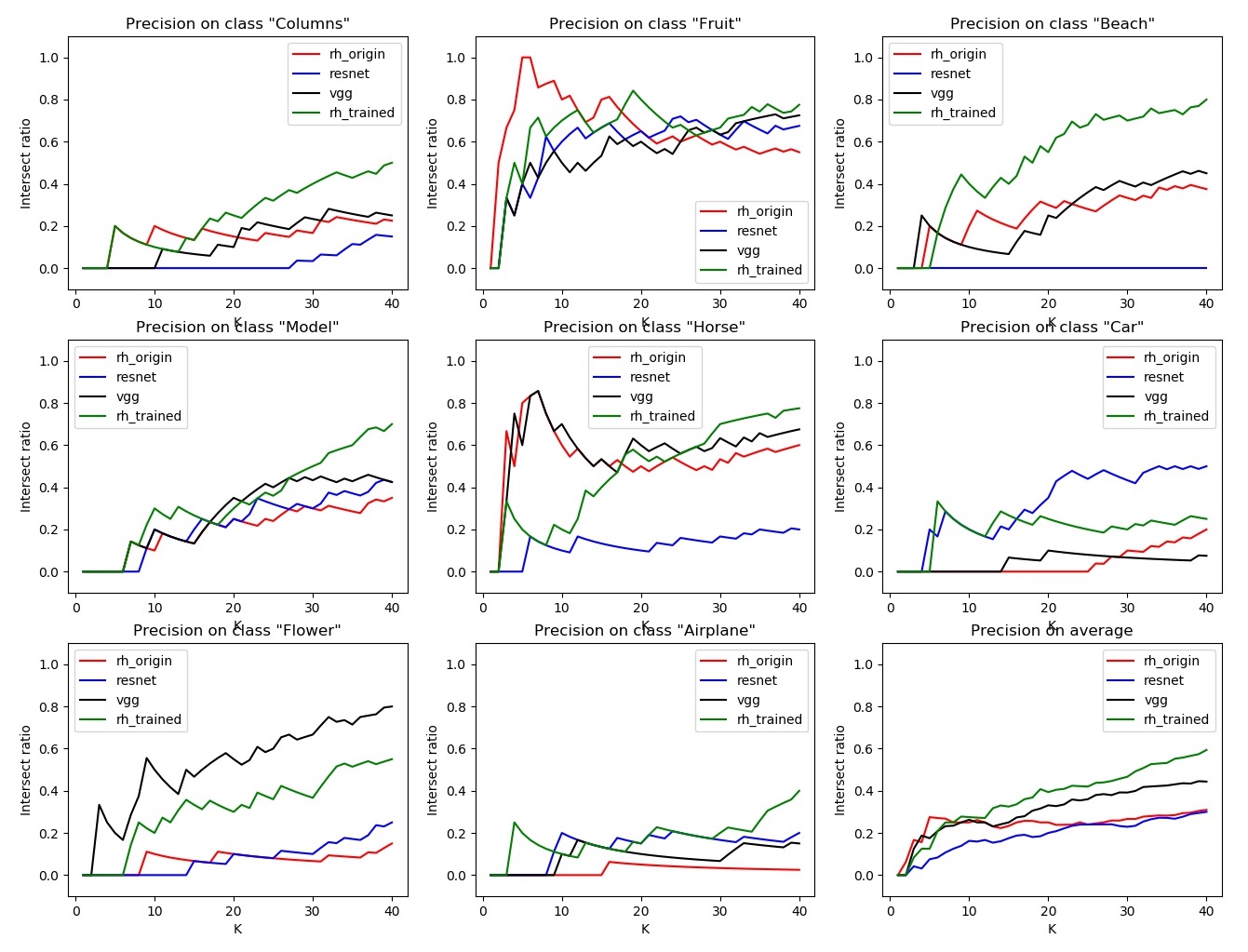} &
\includegraphics[scale=.17]{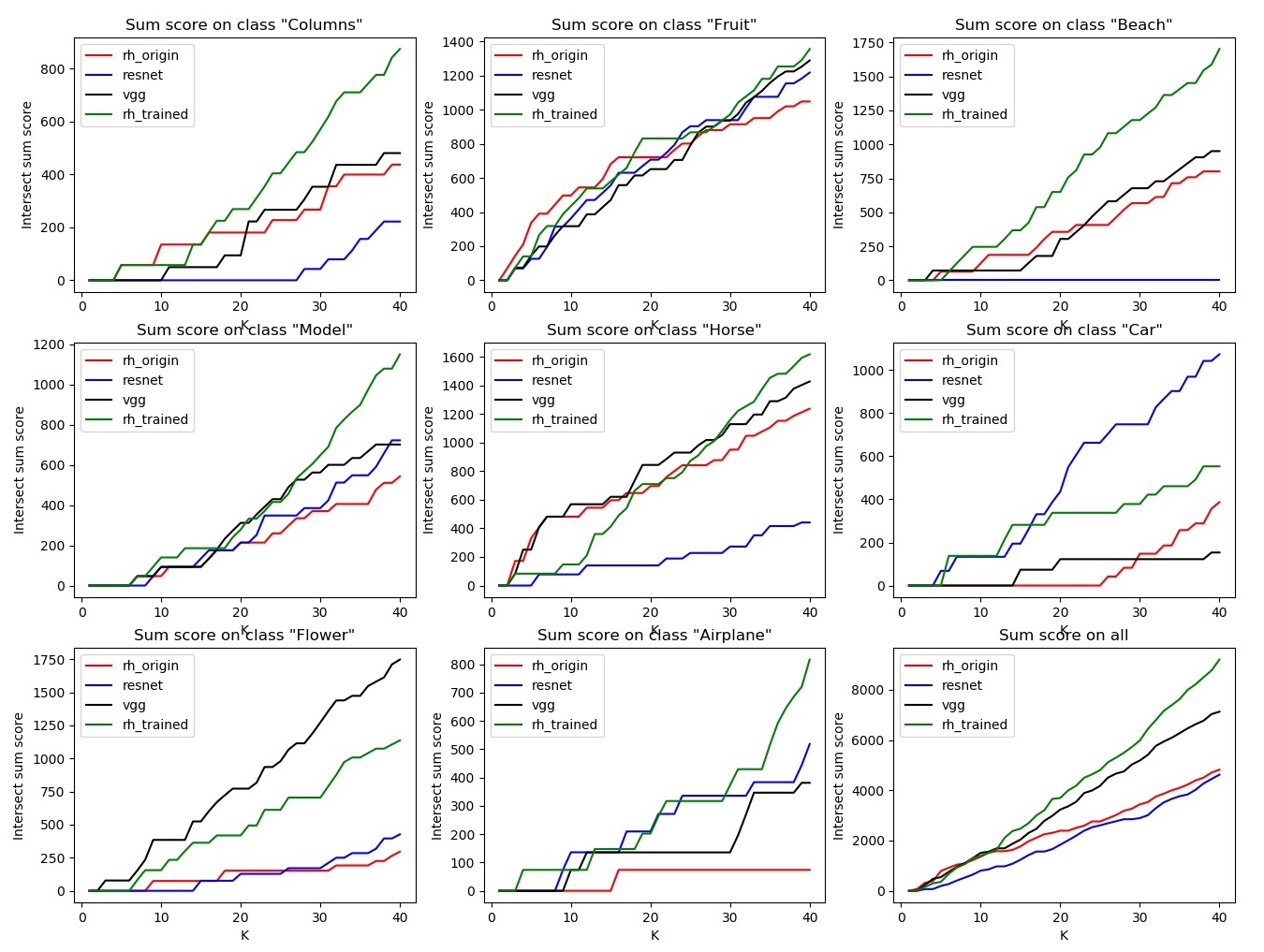} \\
(a) & (b) \\
\end{tabular}
  \caption{Shown here in (a) is the intersection ratio (also known as precision) of top-k images returns in the feature
  space of each trained model. In (b) are the sum of scores of images in the intersection. We denote RDCNN as rh\_origin and RDCNN trained with a single layer neural network (1000 hidden nodes) as rh\_trained. \label{imagesim}}
\end{figure}  

We  obtain the top-k nearest neighbors for each of the eight queries in the four different feature spaces:
fully trained VGG16 and ResNet18, RDCNN, and RDCNN final layer trained with a single layer neural
network.  For each value of $k$ (in $k$-nearest neighbor) we determine the intersection ratio
(also known as the precision) as the number of common images in the true top $k$ images for the
given query divided by $k$. In Figure~\ref{imagesim}(a) we see the intersection ratios for VGG16,
Renset18, RDCNN, and RDCNN followed by a trained single layer neural network.

On the average across the eight queries our trained RDCNN has the highest precision values
as we cross values of $k$ above 10. If we sum the score of all images in the intersection there too we see our trained network in the lead (Figure~\ref{imagesim}(b)). Our random one (without training) is behind VGG but better than 
ResNet. 

While our method performs well for image similarity on this benchmark we see that the fully trained networks 
produce a better classification of the eight queries. 
We consider the correct classification of the query to be the category it belongs to in the database
images. Both VGG16 and ResNet18 classify 7 out of 8 correctly whereas our RDCNN (final trained layer) does 6 out of 8
correctly. On the training VGG and ResNet achieve 98\% accuracy whereas our trained RDCNN gets to 76\%
with the multilayer perceptron. Thus it appears that RDCNN captures image similarity better than image content.

\subsection{Sensitivity to adversarial attacks}
We explore whether the sign activation and lack of gradient in our model offers a defense to adversarial attacks. We perform a practical black box attack \cite{papernot2017practical} on our model and the other trained ones. For each model including ours we first input and output examples from the STL10 dataset. We then use these examples to learn a two layer multi-layer perceptron from which we produce 5,000 adversarial examples. In Table~\ref{adversarial} we see that all models including ours fail on the adversarial examples. This despite the sign activation function and lack of gradient our model can be attacked.

\begin{table}[h]
  \centering
 \setlength\tabcolsep{8pt} 
  \begin{tabular}{ccccc}
  \hline
             & RDCNN & VGG16 & DenseNet40 & ResNet18  \\ \hline
  STL10 & 51\% & 58\% & 46\% & 56\% \\ \hline
  \end{tabular}
  \caption{Accuracy of our model and others on STL10 adversarial examples generated with a black box attack \label{adversarial}}
\end{table}

\section{Discussion}
\begin{figure}[h]
  \centering
  \setlength\tabcolsep{8pt} 
  \begin{tabular}{cc}
\includegraphics[scale=.07]{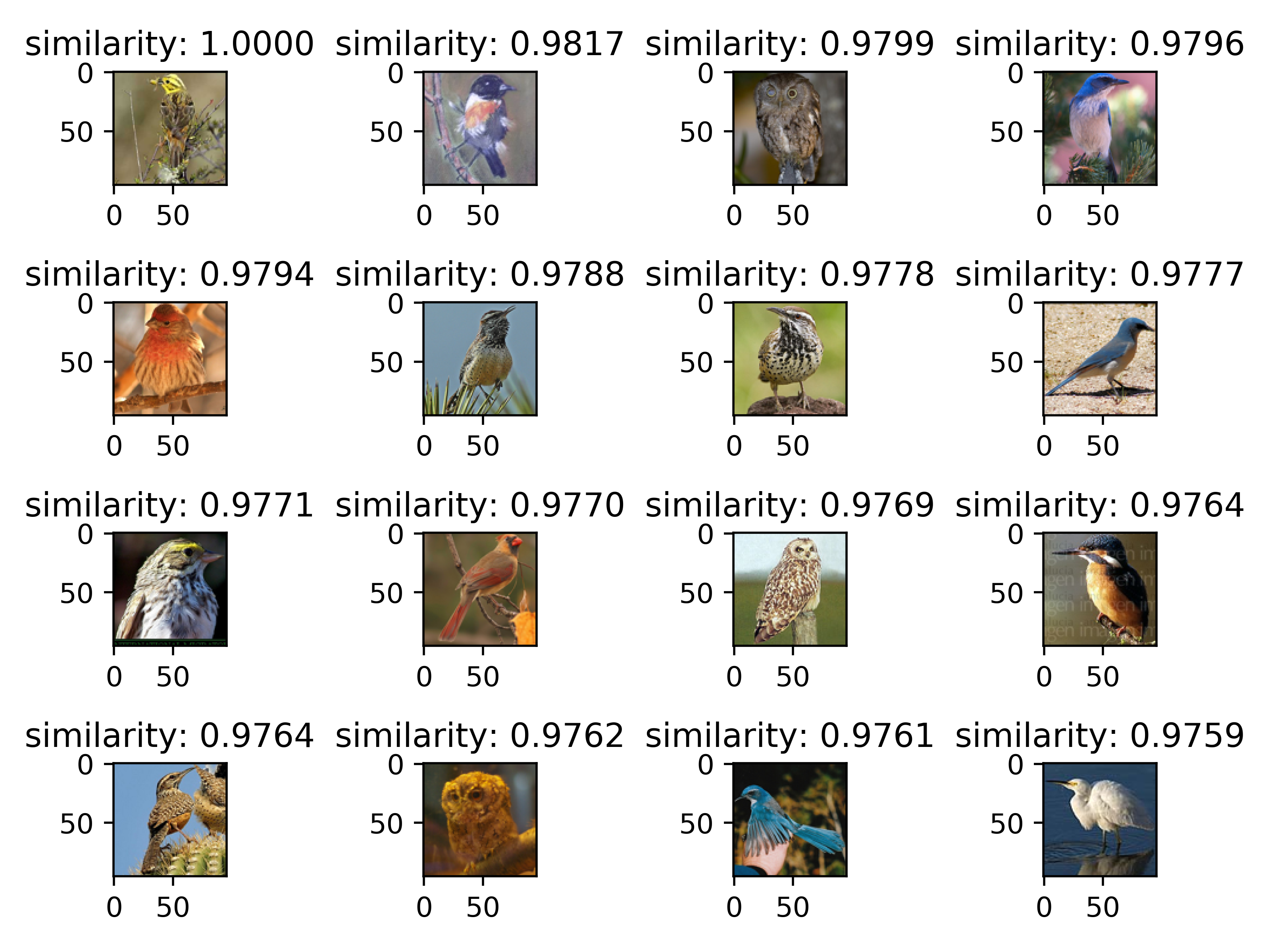} &
\includegraphics[scale=.07]{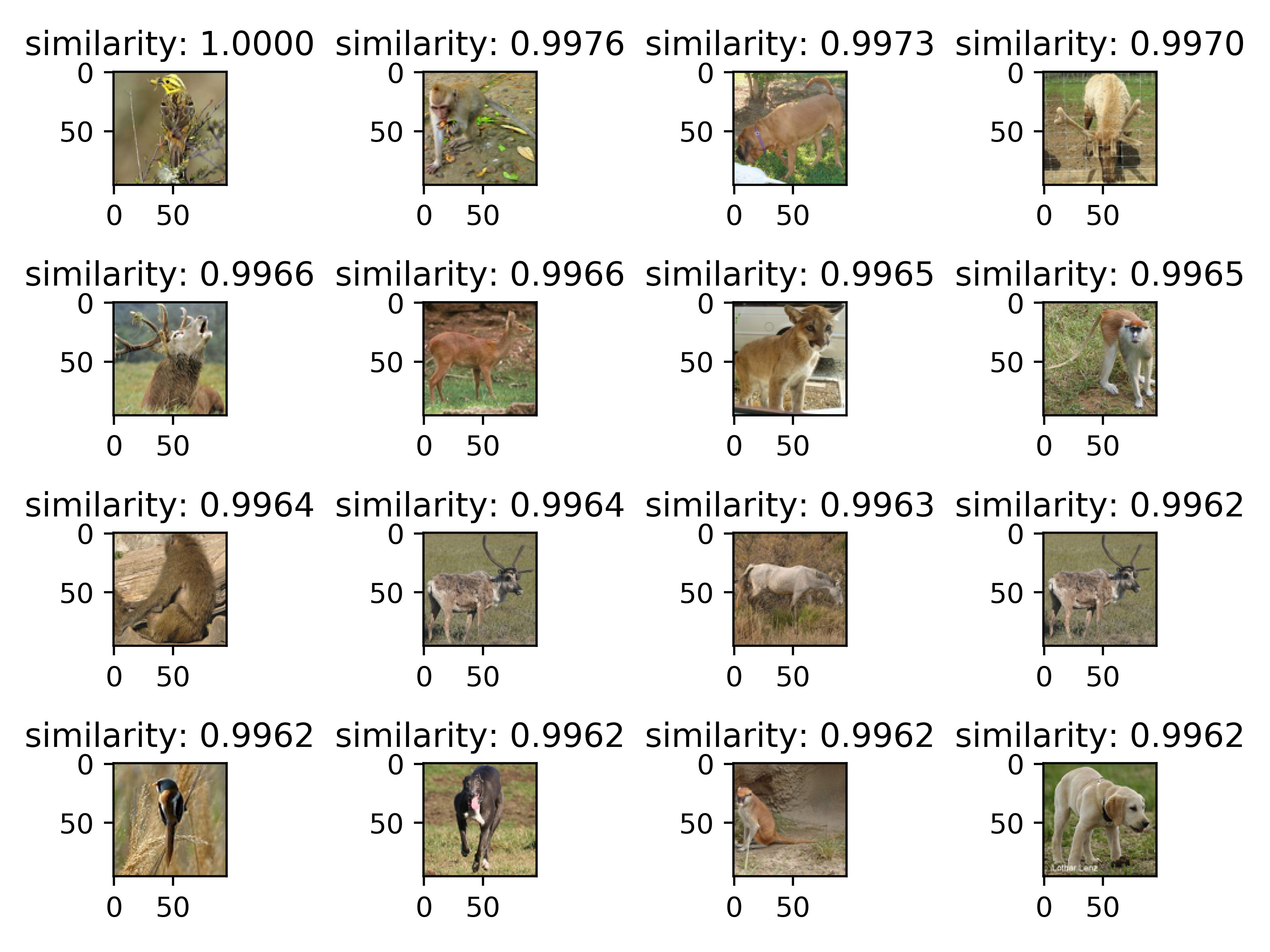} \\
(a) & (b) \\ 
\end{tabular}
\includegraphics[scale=.07]{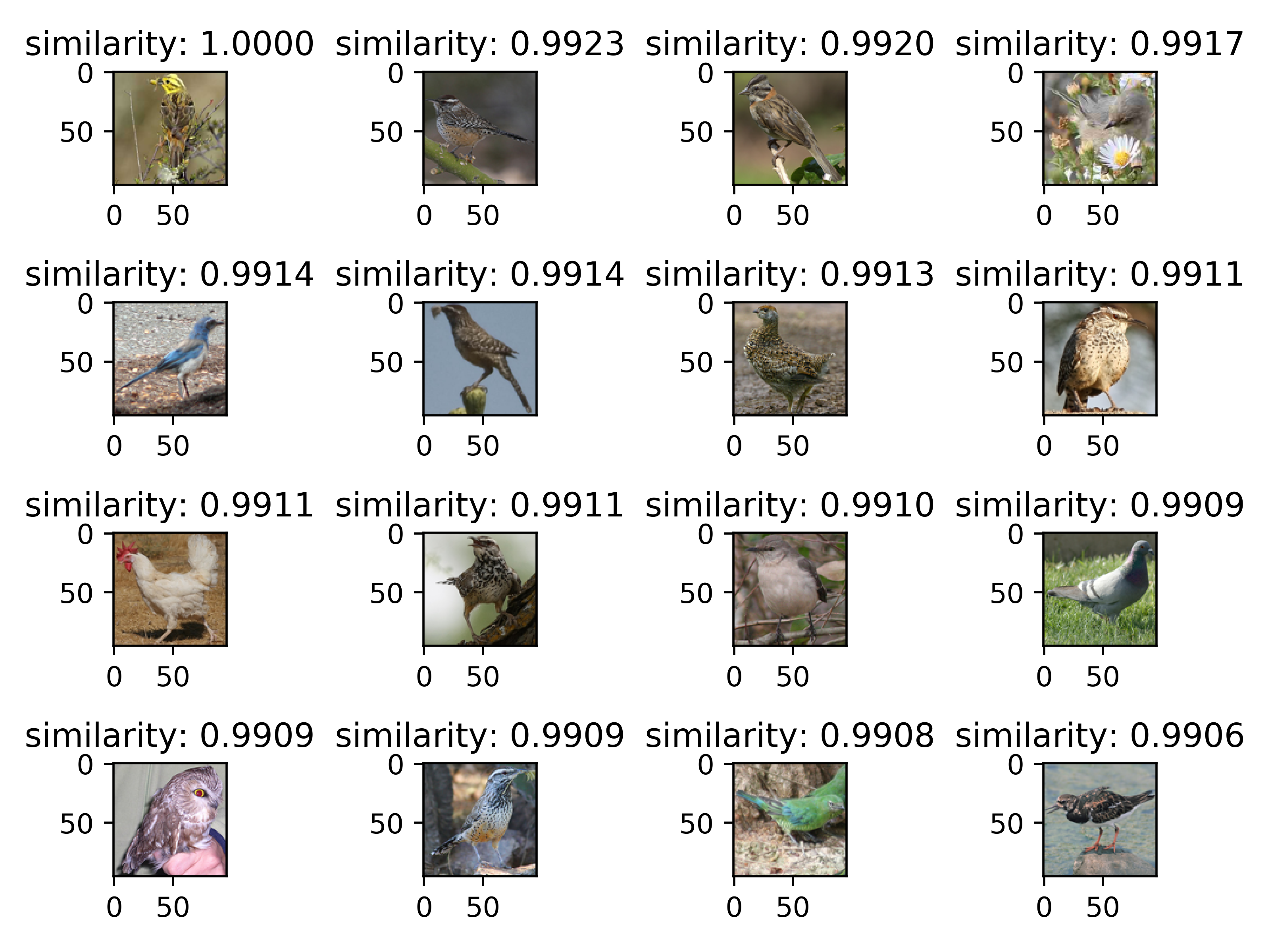} \\
(c)
  \caption{Shown here are cosine similarity values of the top 16 images similar to the bird in the upper left.
  In (a) are images from a trained VGG and in (b) are images from RDCNN's untrained final layer. 
  In (c) are images similar to the bird after training the final layer of RDCNN. \label{sim2}}
\end{figure}

We found that our network is somewhat more sensitive to background than trained networks. From STL10
we pick a bird image and show the top similar images to it as given by VGG16 and our network.
Here we measure cosine similarity in final layer of VGG and our network. In Figure~\ref{sim2}(a) we see
that VGG16 reports almost all birds as the most similar image to the original bird. In our network shown in 
Figure~\ref{sim2}(b) we see that images similar to the bird are also similar in color and background and are mostly not birds.

Thus our random network is sensitive to background and color which possibly accounts for its trailing accuracy in
classification tasks. This can also be seen in the image retrieval experiments where RDCNN performs better in retrieval but worse when we classify the queries. While this obviously is a pitfall it may also be advantageous in problems where similarity plays an important role such as medical imaging \cite{shapiro2008similarity,town2013content}. If we train our network's final layer we begin to see birds in top similar images as shown in Figure~\ref{sim2}(c). Even in the image retrieval by similarity task RDCNN performs better with some training. 

To determine the sensitivity of our method to training set size we reversed our Mini ImageNet training
and test: we use the much smaller test set of 500 images as training and 12,370 for test. In this experiment
we found our method to give 58\% accuracy while a trained ResNet18 gives 48\% and 63.7\% without
and with data augmentation respectively. 

One challenge in RDCNN is the high dimensional feature space. This can be a challenge for large datasets like 
ImageNet \cite{ILSVRC15}. We propose to solve this by compressing the final layer with a 
simple bit-wise encoding method. We plan to study this and applications on medical images as part of future work.


\section{Conclusion}
We propose a random depthwise convolutional neural network with thousands of convolutional kernels per layer.
We show that our network can achieve better accuracies than other random networks and unsupervised feature learning methods, and competitive accuracies in image classification and retrieval compared to trained networks. We highlight sensitivity to background which is a pitfall but also a potential advantage that remains to be explored.


\bibliographystyle{unsrt}
\bibliography{my_bib}




\end{document}